\documentclass[sigconf]{acmart}

\usepackage{subcaption}

\AtBeginDocument{%
  \providecommand\BibTeX{{%
    \normalfont B\kern-0.5em{\scshape i\kern-0.25em b}\kern-0.8em\TeX}}}

\copyrightyear{2020} 
\acmYear{2020} 
\setcopyright{acmcopyright}
\acmConference[IUI '20]{25th International Conference on Intelligent User Interfaces}{March 17--20, 2020}{Cagliari, Italy}
\acmBooktitle{25th International Conference on Intelligent User Interfaces (IUI '20), March 17--20, 2020, Cagliari, Italy}
\acmPrice{15.00}
\acmDOI{10.1145/3377325.3377501}
\acmISBN{978-1-4503-7118-6/20/03}

\begin{document}

\title{Trust in AutoML: Exploring Information Needs for Establishing Trust in Automated Machine Learning Systems}

%


 \author{Jaimie Drozdal}
    \email{drozdj3@rpi.edu}
 \author{Gaurav Dass} 
     \email{dassg2@rpi.edu}
 \author{Bingsheng Yao} 
    \email{yaob@rpi.edu}
 \author{Changruo Zhao} 
    \email{zhaoc6@rpi.edu}
 \affiliation{%
   \institution{Rensselaer Polytechnic Institute}}

 \author{Justin Weisz}
 \email{jweisz@us.ibm.com}
 \author{Dakuo Wang}
 \email{dakuo.wang@ibm.com}
 \author{Michael Muller}
 \email{michael_muller@us.ibm.com}
 \affiliation{%
   \institution{IBM Research}}
   
\author{Lin Ju}
 \email{linju@ca.ibm.com}
 \affiliation{%
   \institution{IBM}}
 
 \author{Hui Su}
 \email{huisuibmres@us.ibm.com}
 \affiliation{%
   \institution{IBM Research}
   \institution{Rensselaer Polytechnic Institute}}







\renewcommand{\shortauthors}{J Drozdal et al.}

\begin{abstract}
We explore trust in a relatively new area of data science: Automated Machine Learning (AutoML). In AutoML, AI methods are used to generate and optimize machine learning models by automatically engineering features, selecting models, and optimizing hyperparameters. 
In this paper, we seek to understand what kinds of information influence data scientists' trust in the models produced by AutoML? We operationalize trust as a willingness to deploy a model produced using automated methods. We report results from three studies -- qualitative interviews, a controlled experiment, and a card-sorting task -- to understand the information needs of data scientists for establishing trust in AutoML systems. We find that including transparency features in an AutoML tool increased user trust and understandability in the tool; and out of all proposed features, model performance metrics and visualizations are the most important information to data scientists when establishing their trust with an AutoML tool.

\end{abstract}

\ccsdesc[300]{Human-centered computing~User studies}
\ccsdesc[300]{Human-centered computing~Empirical studies in HCI}
\ccsdesc[300]{Computing methodologies~Artificial intelligence}

\keywords{AutoAI, AutoML, AutoDS, Automated Artificial Intelligence, Automated Machine Learning, Automated Data Science, Trust}


\maketitle

\section{Introduction}
The practice of data science is rapidly becoming automated. New techniques developed by the artificial intelligence and machine learning communities are able to perform data science work such as selecting models, engineering features, and tuning hyperparameters~\cite{khurana2016cognito, kanter2015deep,lam2017one, zoller2019survey,liu2019admm}. Sometimes, these automated methods are able to produce better results than people~\cite{malaika_wang_2019}. Given the current shortage of data scientists in the profession~\cite{dubois2019shortage}, automated techniques hold much promise for either improving the productivity of current data scientists~\cite{wang2019humanai} or replacing them outright~\cite{recio2018can}.

Many companies and open source communities are creating tools and technologies for conducting automated data science ~\cite{web:googleautoml, web:microsoftazure,web:ibmautoai, web:h2o, web:datarobot, web:tpot, olson2016tpot, web:autosklearn}. However, in order for these automated data science techniques -- which we collectively refer to as ``AutoML'' -- to become more widely used in practice, multiple studies have recently suggested that a significant hurdle in \emph{establishing trust} must first be overcome~\cite{wang2019humanai,wang2019atmseer,lee2019human,golovin2017google}. Can AI-generated models be trusted? What factors contribute to the trust of AutoML systems?

In this paper, we discuss users' \textbf{trust} in AutoML systems as it pertains to \textbf{transparency} and \textbf{understandability}, thus we believe it is necessary to clarify these three concepts. 
Our transparency concept derives from  \cite{mercado2016intelligent} that transparency of the automation is ``the quality of an interface pertaining to its ability to afford an operator's comprehension about an intelligent agent's intent, performance, future plans and reasoning process.''
Understandability is the quality of comprehensibility in an automation tool. However, high transparency of a system does not necessarily lead to high understandability.

In this paper, we use a definition of trust as ``the extent to which a user is confident in, and willing to act on the basis of, the recommendations, actions, and decisions of an artificially intelligent decision aid'' \cite{madsen2000measuring}.
Furthermore, we adopt an informational perspective of trust~\cite{zhao2019users, hancock2011meta, siau2018building,hoffman2018metrics}: having information about how an AutoML system works, as well as the artifacts it produces (i.e. machine learning models), ought to increase trust in that system due to increased levels of transparency. Conversely, not having information about an AutoML system or its artifacts ought to decrease trust in that system. Thus, we seek to understand what kinds of information are important for an AutoML system to include in its user interface in order to establish trust with its users. 

Current AutoML systems offer a myriad of information about their own operation and about the artifacts they produce. We coarsely group this information into three categories: information about the \emph{process} of how AutoML works (e.g. information about the \emph{data} (such as how it was transformed or pre-processed), how it performs feature engineering), and information about the \emph{models} produced by the system (such as their evaluation metrics).

In this paper, we report results from three studies designed to evaluate how the inclusion or exclusion of these types of information impacts data scientists' trust in an AutoML system. Our first study is formative, consisting of a number of semi-structured interviews designed to capture the universe of information currently present across a representative sample of commercial AutoML products, and identify information that is not commonly included in those products. We hypothesize that the inclusion of this ``hidden'' information will increase the transparency of AutoML systems, and hence, increase the amount of trust users' place in it. Our second study is evaluative, designed to quantitatively test the impact of the inclusion of new information -- what we refer to as ``transparency features'' -- on ratings of trust. Our third study is an open card-sorting task that aims to understand the relative importance of different kinds of information in the AutoML ``information space.''

We focus on addressing two research questions in our work.

\begin{itemize}
    \item \textbf{RQ1}. To what extent does the inclusion of new transparency features affect trust in and understanding of an AutoML system?
    \item \textbf{RQ2}. What information is highly important for establishing trust in an AutoML system? What information is not important?
\end{itemize}

Our results make a number of significant contributions to the existing literature on data science work and to the IUI community.

\begin{itemize}
    \item We find quantitative evidence that the inclusion of transparency features -- visualizations of input data distributions and a visual depiction of the feature engineering process -- increases peoples' ratings of trust and understanding of an AutoML system.
    \item We provide a ranking of relative importance of different kinds of informational ``nuggets'' in establishing trust in an AutoML system.
\end{itemize}

We expect our work to inform the design of AutoML systems by highlighting the different types of informational needs data scientists have in order to establish trust in the system.


\section{Related Work}
We first review literature on human-in-the-loop machine learning, focused on understanding the work practices and tool use of data scientists. We then discuss recent advances in automated data science, and summarize issues of trust and transparency in machine learning.


\subsection{Human-in-the-loop Machine Learning}
Data science is the process of generating insights from primarily quantitative data~\cite{kross2019practitioners}. Often, data scientists leverage techniques from machine learning to build models that make predictions or recommendations based on historical data. Studies have suggested that data science work practices are different from traditional engineering work~\cite{kross2019practitioners, guo2011proactive, muller}. For example, Muller et al.~\cite{muller} decomposed the data science workflow into 4 sub-stages, based on interviews with professional data scientists: data acquisition, data cleaning, feature engineering, and model building and selection. They argued that a data science task is more similar to a crafting work practice than an engineering work practice, as data scientists need to be deeply involved in the curation and the design of data, similar to how artists craft their work.

\begin{figure}[ht]
  \begin{center}
    \includegraphics[width=\columnwidth]{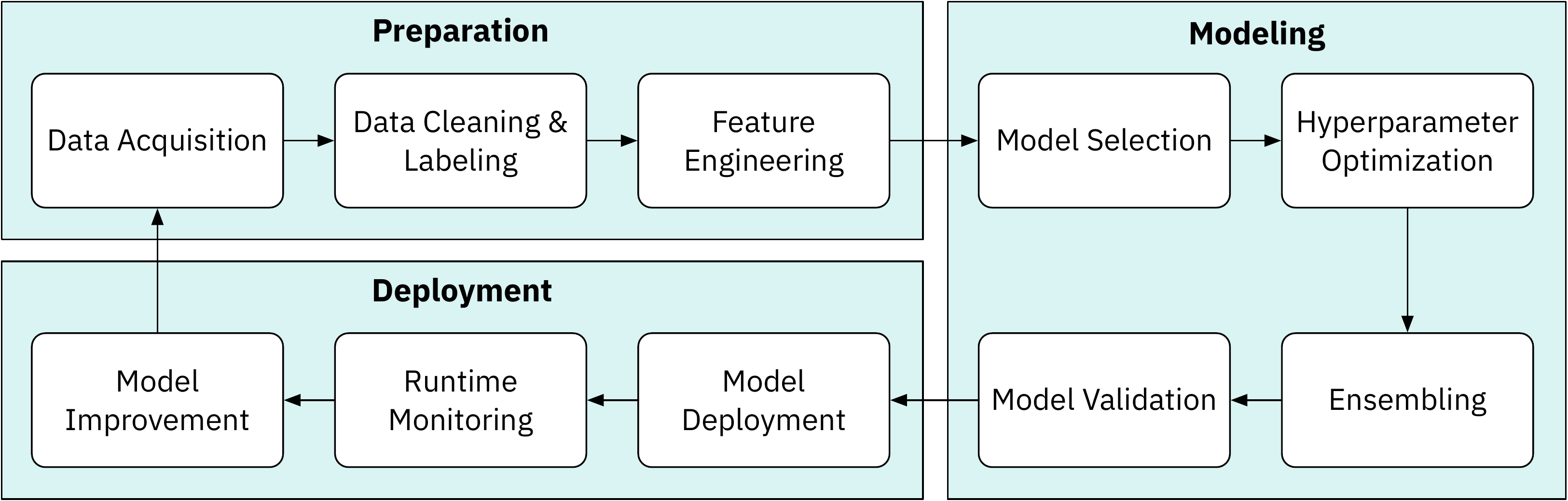}
  \end{center}
  \caption{A data science workflow, consisting of three high-level phases: data preparation, model building, and model deployment ~\cite{wang2019humanai} }
  \label{fig:DS-steps}
\end{figure}

Wang et al. proposed a three stage framework of data science workflow with ten sub-steps~\cite{wang2019humanai}. It expands ~\cite{muller} workflow, which mostly focuses on the model training steps and takes into account the model deployment steps after the model is trained. In this paper, we adopt the framework of \cite{wang2019humanai}, as shown in Figure ~\ref{fig:DS-steps}, and we focus mostly on data scientists' trust in the \emph{model validation} sub-step of the \emph{modeling} phase.

CSCW researchers have also looked at collaborative aspects of data science work. Hou and Wang ~\cite{hou2017hacking} conducted an ethnography study to explore collaboration in a civic data hackathon event where data science workers help Non-profit organizations to develop insights from their data. Mao et al.~\cite{mao2019} interviewed biomedical domain experts and data scientists who worked on the same data science projects. Their findings partially echo previous literature~\cite{muller,guo2011proactive,kross2019practitioners} results that data science workflow has multiple steps. In addition, they also suggest that data science is a highly collaborative effort where domain experts and data science workers need to work closely together to advance the workflow. Often times, the two parties are not ``speaking the same language'' ~\cite{hou2017hacking} or do not have the common ground related to their goal~\cite{mao2019}. Thus, a ``broker role'' in the team, who can bridge the two backgrounds and constantly re-calibrate the goal, may help to ease these tensions and support the success of this cross-discipline data science work.

These empirical findings guided the designers and system builders to propose human-in-the-loop data science design principles~\cite{amershi2019guidelines,amershi2011human,wang2019data,kery2018story,kery2019towards,gil2019towards}. Gil et al. surveyed papers about building machine learning systems and developed a set of design guidelines for building human-centered machine learning systems~\cite{gil2019towards}. Amershi et al. in parallel reviewed a broader spectrum of AI applications and proposed a set of design suggestions for AI system in general~\cite{amershi2019guidelines}, some of which are overlapping with the ones in ~\cite{gil2019towards}. 

With these design suggestions, more and more machine learning tools are built to support data scientists. For example, Jupyter Notebook is one of the successful examples~\cite{web:jupyter}. It incorporates data scientists' three needs of coding, documenting narrative, and observing execution results~\cite{kross2019practitioners} into a cohesive user interface. Thus, many data science workers adopt Jupyter notebooks as their primary working environment. Researchers have studied how data scientists use Notebooks~\cite{rule2018exploration,passi2018trust}, how to build version control components into Notebooks~\cite{kery2019towards}, how to enable multi-users synchronous editing in Notebook~\cite{wang2019data}, and other innovative designs. 

In this paper, we build upon the understandings of how data science teams work and how to build systems to support data scientists work practice, but focus specifically on the scenario where users interact with automated machine learning techniques in a data science work practice.

\subsection{Automated Machine Learning (AutoML)}

Automated Data Science or Machine Learning (AutoML for short) refers to systems that automatically select and optimize the machine learning model in each of the data science steps~\cite{wang2019atmseer}. For example, AI researchers have developed various algorithms to automatically \emph{clean data}~\cite{kougka2018many}. The \emph{feature engineering} step is tedious for human data scientists thus lots of solutions have been proposed to automatically generate new features and select the best subset of the features while balancing good model performance~\cite{kaggle2018survey,lam2017one,kanter2015deep}. As for the \emph{model selection and hyperparameter optimization} steps, data scientists are already relying on publicly available libraries such as Auto-sklearn~\cite{web:autosklearn} and TPOT~\cite{web:tpot}, instead of writing code from scratch.

However, all of these automation technologies focus only on a single step and the data scientists still need to put various pieces of the puzzle together to assemble a model generation pipeline. End-to-end AutoML solutions have only recently become a reality. These technologies arguably can complete the entire data science workflow (as in Figure~\ref{fig:DS-steps}) from data acquisition to model selection, and then to model deployment and improvement. Large technology companies have released automated data science products, such as Google's AutoML~\cite{web:googleautoml}, IBM's AutoAI~\cite{web:ibmautoai}, and Microsoft's Azure Studio~\cite{web:microsoftazure}. Small startups such as H2O~\cite{web:h2o} and Data Robot~\cite{web:datarobot} are also capturing significant market share.

With these latest developments, Automated Machine Learning becomes increasingly promising, and more and more researchers have begun to explore its user experience~\cite{lee2019human,wang2019humanai, gil2019towards}. For example, Gil et al. reviewed existing literature on how data scientists use machine learning applications, from which they proposed design guidelines for the development of future AutoML systems~\cite{gil2019towards}. Wang et al. interviewed 20 professional data scientists and asked their perceptions on AutoML technology ~\cite{wang2019humanai}. They found that data scientists in general hold a positive position towards the collaborative future where ``human and AI work together to build models''. Lee et al. referenced back to the ``mixed-initiative'' literature and argued that AutoML and human users can "collaborate efficiently to achieve [the] user's goals"~\cite{lee2019human}, thus it is a human-in-the-loop perspective of AutoML.

This emerging group of empirical studies enrich our understanding of how data scientists think, interact, and collaborate with AutoML tools and generate useful design guidelines to improve an AutoML system's usability. However, as recent works suggested~\cite{wang2019atmseer,golovin2017google,park2019visualhypertuner,lee2019human}, transparency and trust in this new AutoML technology is another major hindrance for large scale adoption. Thus, in the next subsection, we will focus on the transparency and trust issues of AutoML.

\subsection{Trust in Machine Learning and in AutoML}
AutoML is a relatively new topic, and therefore not many works have investigated trust issues of AutoML systems in particular - e.g., ~\cite{wang2019humanai,lee2019human,gil2019towards}. All of these works made design suggestions that, in order to make AutoML systems accessible and effortless for the data scientist users in the future, designers and system builders should present not only the final model results coming out of the system, but also the pipeline steps and decisions made in each of those steps along with the model generation process. This argument implies users demand higher transparency from AutoML system.

To accommodate the user needs of transparency and trust, a few recent works proposed various design prototypes for increasing AutoML systems transparency ~\cite{wang2019atmseer,golovin2017google,park2019visualhypertuner,lee2019human}. For example, Wang et al. developed a first of its kind visualization systems, ATMSeer, that aims to open up the blackbox of an end-to-end AutoML system~\cite{wang2019atmseer}. The authors argue that their tool can provide a multi-granularity visualization (both for the model selection as well as for hyperparameter selection) to the users so that they can monitor the AutoML process and adjust the search space in real time \cite{wang2019atmseer}.

This approach of using visualizations to increase transparency of a system has been common in the traditional machine learning system designs~\cite{dakuo,Wang:2015:DVC:2702123.2702517,DBLP:journals/corr/abs-1906-07716}. For example, Google Vizer is a visualization tool that can reveal the optimization details of the hyperparameter tuning step ~\cite{golovin2017google}. In addition, it provides a visualization that shows the range of each hyperparameter in a model and the relationship between performance and hyperparameters. The authors hope that in this way, the users can understand how a final choice of  the hyperparameter value is decided among alternative options (i.e., it leads to better model performance). VisualHyperTuner is a similar visualization-based system that focuses only on the hyperparameter tuning step \cite{park2019visualhypertuner}.

All of these systems fall short in presenting an overview of the AutoML process and how each model pipeline was created. Dignum et al. suggests that transparency resides in not only the result of the model but also the data and the processes where the model has been generated~\cite{dignum2017responsible}. This task is challenging for a visualization, as some of the steps in the AutoML pipeline have a categorical search space (e.g., various algorithms in the model selection step) but others have a continuous search dimension (e.g., hyperparameters values in the optimization step). Furthermore, among these AutoML steps, some have a sub-step or even a sub-sub-step. Thus, using one visualization to convey both the overview view of the pipeline as well as the necessary details of each step in the workflow is challenging. 

ATMSeer~\cite{wang2019atmseer} presents a nice framework of proposing a visualization prototype on top of an existing AutoML system, and evaluating it with a number of users in a user study. Our paper adopted this research methodology and we propose low-fidelity design prototypes (based on the feedback from a Think-Aloud study) and evaluate these features with users on the perceived transparency level of AutoML. We hope these features can increase AutoML transparency and further promote better collaboration and trust between the data scientist and AutoML during the \emph{Preparation and Modeling} phases (Figure \ref{fig:DS-steps}). We leave the transparency topic in the \emph{Deployment} phase for future work.

In particular, we join the group of researchers who argue that higher transparency of an AI system leads to higher trust by a user of that system~\cite{hancock2011meta,zhao2019users,siau2018building, liao2020questioning,zhang2020effect}. Because transparency and trust of AutoML systems is a relatively new research topic, we want to build our baseline understanding about what information users need and how they need those information from an AutoML user interface. We designed a think-aloud pilot study as our first of a series of three studies. Based on the findings from study 1 with four data science students, we operationalize the transparency of an AutoML system into various design features such as ``showing the distribution of data column (data-oriented transparency)'' or ``showing how AutoML performs feature engineering (process-oriented transparency)''. We hope a comparison user study (study 2) between users who are and those who are not exposed to these transparency features can reveal the quantifiable differences in their trust in the AutoML system. Throwing more information to users can always increase the transparency of AutoML, but we are in danger of users' cognitive overload. To prioritize users information needs, we designed a study 3 that asks users to do a card-sorting task to prioritize those needs.

In what follows, we will start reporting our study designs and findings of the three user studies in order.

\section{Study 1: Think-Aloud Evaluation} 
\label{sec:study1}
We conducted a small pilot study with four computer science Graduate students to understand the current ``information landscape'' of AutoML systems. Participants were asked to use four popular commercial AutoML products while thinking aloud about the tool's information design and how it affected their feelings of trust in the tool.

Participants were presented with four tools, in random order: Google's AutoML Tables~\cite{web:googleautoml}, Microsoft's Azure Machine Learning Studio~\cite{web:microsoftazure}, IBM's Watson Studio AutoAI~\cite{web:ibmautoai}, and H2O's Driverless AI~\cite{web:h2o}. We selected these tools based on their popularity, the fact that they require no coding from the user, and the fact that they automate the entire machine learning workflow from data preparation to model evaluation. Participants were given 30 minutes with each tool, with the task of generating and evaluating models for the Titanic dataset~\cite{web:titanic}. 

Participants were asked to think aloud as they interacted with each tool. Participants were also prompted with questions such as, ``is this tool showing you everything you need at this moment?'' and ``how is this [feature/visualization/information] affecting your trust in the tool?'' At the end of reviewing all four tools, participants were asked which of the four tools they preferred.


Our procedures were approved by our institution's review board, and participants provided written informed consent before participating. Participants were not compensated for their participation in this study.

\subsection{Results}
Overall, transparency was a highly-desired feature from all participants. Specifically, transparency of both \emph{data} and \emph{models} were mentioned while interacting with all four of the tools.

Participants' feedback was also varied, with many individual differences in preferences. Each of the four participants preferred a different tool. Two participants were hyper-focused on information about data and consistently commented on the lack of support for having conversations with the data~\cite{muller}, either in raw or pre-processed form. Another participant cared the most about the model selection process and noted the lack of transparency around the set of candidate models that the tools considered when performing model selection.

All participants expressed a lack of understanding about the different processes used by the AutoML tools.

\begin{quote}
    \textit{``I have no idea what was done to the data.'' (P2)}
\end{quote}

\begin{quote}
    \textit{``I would not use this software because it's not clear what is happening when the experiment is running.'' (P3)}
\end{quote}

\begin{quote}
    \textit{``Feature engineering is complicated and important, but I don't know how it's doing it.'' (P4)}
\end{quote}

Based on feedback from participants in this study, we focused on understanding how two transparency features might affect trust in an AutoML tool: visualizations of data distributions, and a visual depiction of how the feature engineering process works. We explored the effect of these transparency features in Study 2, discussed in the next section. Additional feedback from this study regarding the different kinds of information present in each of the AutoML interfaces was used in the design of Study 3, discussed in Section~\ref{sec:study3}.

\section{Study 2: Increasing Trust via Transparency Features}
\label{sec:study2}

Study 1 suggested that commercial AutoML systems at the time of the study were lacking in transparency. In this study, we conducted a more detailed examination of how increasing transparency would affect trust. We recognize that transparency may be provided at different levels:

\begin{itemize}
    \item data-oriented transparency (e.g. showing data distributions for the columns in the training set)
    \item process-oriented transparency (e.g. how AutoML performs feature engineering or hyperparameter optimization)
    \item model-oriented transparency (e.g. showing various accuracy metrics on a validation set)
\end{itemize}

In this study, we compared a baseline AutoML user interface with one that included additional \emph{transparency features} that provided additional insight into the distributions of input features (data-oriented transparency) and the process by which the AutoML engineers new features (process-oriented transparency). These transparency features were chosen due to the overwhelming amount of feedback from participants in Study 1, who felt that both of these features were important, yet missing from the AutoML systems. They may appear simple, but at the time of the study these features did not exist in a majority of AutoML tools. Not to mention, while including distributions of input features as a transparency feature might be argued as overly simplified the future of AutoML is one in which the entire data science process is captured in a single environment. 

\begin{figure*}[htp]
    \centering
    \subcaptionbox{Example of input data distribution for one of the columns in the data set. Written explanations accompanied each distribution to provide additional insight in interpreting the chart, as a proxy for the subject matter expertise typically available when conducting data science work.} {
        \frame{\includegraphics[height=6.5cm]{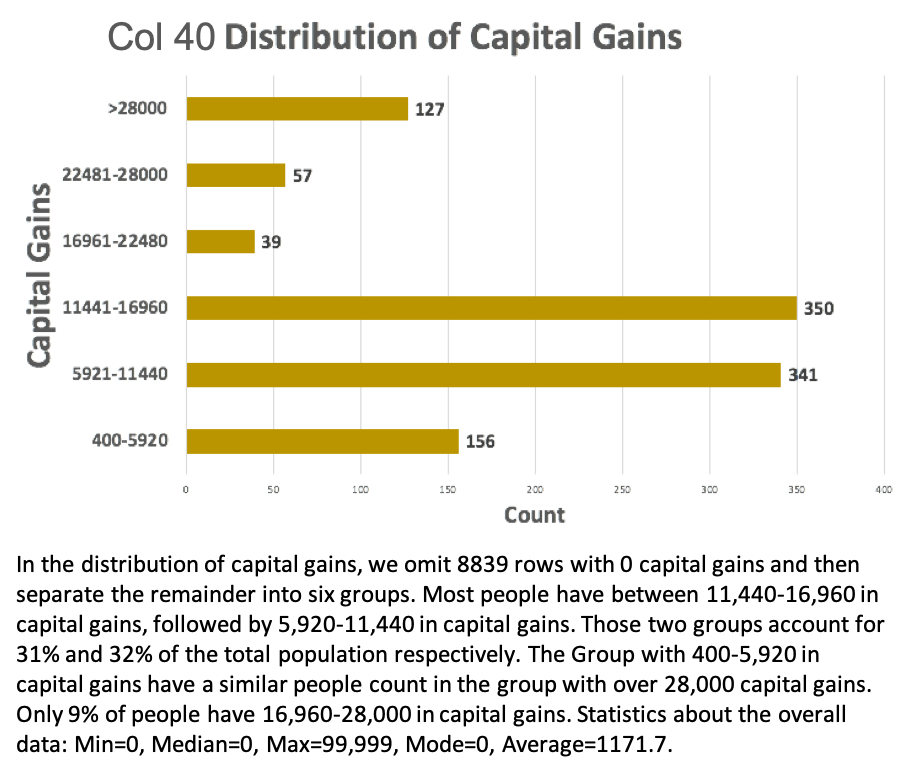}}
    }
    \hspace*{\fill} 
    \subcaptionbox{Diagram illustrating the feature engineering process. Nodes represent data set versions and lines represent transformations applied to one of the data set's features, producing a new ``view'' of the data set. $D_0$ is the base data set with no transformations applied. At each step, accuracy is evaluated to determine if the transformed features provide an increase in model performance. The blue highlight represents how a sequence of feature transformations results in a model with a higher accuracy.\cite{khurana2016cognito}\cite{khurana2018feature}}{
        \frame{\includegraphics[height=6.5cm]{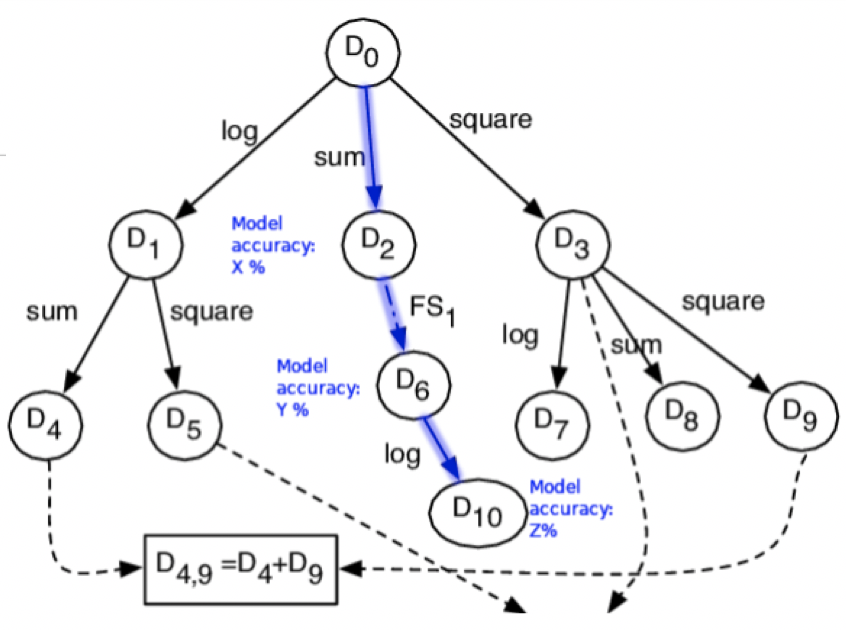}}
    }
    \caption{(a) Input data distribution transparency feature. (b) Feature engineering transparency feature.}
    \label{fig:transparency}
\end{figure*}

To evaluate these transparency features, we used screenshots from a commercially-available AutoML system, removing references to the company's name and logo, in order to provide both a realistic AutoML experience, and one in which we could easily incorporate the transparency features. Specifically, we used IBM's Watson Studio AutoAI~\cite{web:ibmautoai}. Figure~\ref{fig:transparency} shows examples of how these transparency features were shown to participants.  

\subsection{Participants}
We recruited 21 participants who had prior experience with machine learning to complete our study. One participant was dropped from our study due to a lack of knowledge about machine learning that was uncovered during the course of the study; thus, our final sample consists of $N=20$ participants.

Of the 20 participants, 5 (25\%) were Undergraduate students and 15 (75\%) were Graduate students. Five participants (25\%) were female and 15 were male (75\%), which is slightly higher than the proportion of women in data science (16.8\%) reported in the 2018 Kaggle data science survey~\cite{kaggle2018survey}. Students' areas of study included information technology (35\%), computer science (20\%), business analytics (15\%), mathematics (10\%), quantitative finance (10\%), and other disciplines (10\%).

\subsection{Procedure}
Prior to participating in our study, participants listened to the nature of the study and its procedures. Participants provided written informed consent. Our study was reviewed and approved by the institutional review board at our institution.

Participation in the study took approximately one hour, and participants received a \$10 USD gift card for their time.\footnote{The local minimum wage at the time of study was \$9.70.} We began by giving participants a packet of information about the dataset, which included column meanings, and all of the documentation on the tool available online from the provider. These handouts were to be used as reference materials while completing the study. Next, the experiment proceeded in two phases -- reviewing AutoAI-produced models (described in the next section) and a card-sorting task (described in Section~\ref{sec:study3}).

\subsubsection{Reviewing AutoML-produced models}
In this phase, participants completed a sequence of two tasks. Each task consisted of reviewing a packet~\footnote{We opted not to run the actual AutoAI system during the course of the study as it would take too much time. Therefore, we ran the system on the data on ourselves and compiled a series of screenshots from the tool's user interface for participants to review.} of information about an AutoAI ``run'' for the Census-Income dataset~\cite{Dua:2019}. This dataset is used to train a model that predicts whether a loan application should be approved from a set of over 35 different factors.

In this run, AutoAI produced a set of four pipelines. Each pipeline consisted of a series of steps, such as model selection, feature engineering, and hyperparameter optimization.\footnote{We use the terms ``pipeline'' and ``model'' interchangeably to refer to the resulting output of the AutoAI process: a binary classifier for making loan approval decisions. Some, but not all pipelines generated by AutoAI include feature engineering and hyperparameter optimization.} Participants reviewed a number of details about these pipelines, such as performance metrics and confusion matrices. Participants were asked by the researcher whether they trusted any of the four models enough that they would use them in a real deployment. After this, participants filled out a questionnaire containing demographic questions and different Likert scale questions measuring trust and understandability described in the following section. 

To compare different user experiences of AutoAI, we developed three separate packets of information. The ``V1'' packet represents the base user interface provided by the commercial AutoML system we used. From this, we developed two variants, ``V2A'' and ``V2B,'' by adding in additional visualizations of the input data and feature engineering process, respectively. We outline the informational content of each packet in Table~\ref{tab:info_packet}. In the study, all participants saw the V1 packet, but each participant saw only one of the V2A or V2B packets. Packets were shown in a random order to control for order effects, and equal numbers of participants saw the V1 and V2 packets first.

\begin{table}[ht]
    \small
    \centering
    \begin{tabular}{p{5.5cm}|c|c|c}
         \textbf{Information}       & \textbf{V1} & \textbf{V2A} & \textbf{V2B} \\
         \hline
         \textbf{Data} & & & \\
         \hspace{0.1cm} Raw data table.   & \checkmark & \checkmark & \checkmark \\
         \hspace{0.1cm} Charts of input feature distributions & & \checkmark & \\
         \textbf{Process} & & & \\
         \hspace{0.1cm} Visualization of pipeline creation process & \checkmark & \checkmark & \checkmark \\
         \hspace{0.1cm} Feature engineering process diagram & & & \checkmark \\
         \textbf{Model} & & & \\
         \hspace{0.1cm} Metrics (ROC AUC, accuracy, $F_1$, etc.) & \checkmark & \checkmark & \checkmark \\
         \hspace{0.1cm} ROC curve & \checkmark & \checkmark & \checkmark \\
         \hspace{0.1cm} Precision Recall curve & \checkmark & \checkmark & \checkmark \\
         \hspace{0.1cm} Confusion matrix & \checkmark & \checkmark & \checkmark \\
         \hspace{0.1cm} Feature importance chart & \checkmark & \checkmark & \checkmark \\
         \hspace{0.1cm} \hangindent=0.3cm Feature transformation table (for pipelines that included feature engineering) & \checkmark & \checkmark & \checkmark \\
         \hline
    \end{tabular}
    \caption{Information included in each of the V1, V2A, and V2B packets. All participants saw the V1 packet, but only one of the V2A and V2B packets. The packets were presented in random order.}
    \label{tab:info_packet}
\end{table}

\subsection{Measures}
For each of the phase 1 packets, we asked participants whether they trusted any of the pipelines produced by AutoAI enough to use them in a real deployment of an AI system. In addition, we asked participants to rate their overall trust in the AutoML system, based on the Merritt scale~\cite{merritt2011affective} reported in Hoffman et al.~\cite{hoffman2018metrics}. This scale treats trust as an attitudinal judgement of the degree to which a person can rely on an automated system to achieve their goals. We included 4 items from this scale related to confidence and dependability, and adapted them to our particular situation: ``I believe the tool is a competent performer,'' ``I trust the tool,'' ``I have confidence in the advice given by the tool,'' and ``I can depend on the tool.'' These items were rated on 5-point Likert scales (Strongly Disagree to Strongly Agree).

Another aspect of trust has to do with understandability~\cite{madsen2000measuring, hoffman2018metrics}. We developed a 9-item scale to assess the degree to which participants understood the AutoML tool presented in each packet. Whereas the Madsen-Gregor scale~\cite{madsen2000measuring} for evaluating perceived understandably contains high-level statements applied to a whole system (e.g. ``I understand how the system will assist me with decisions I have to make''), we desired a scale that included items specifically focused on aspects of AutoML systems. Our scale included the following items: ``I understand the tool,'' ``I understood the tool's overall process,'' ``I understand the data,'' ``I understand how the tool performs data preprocessing,'' ``I understand how estimators are selected,'' ``I understand how new features are generated by the tool,'' ``I understand the differences between the generated models,'' ``I understand the model evaluation metrics,'' and ``I understand the model evaluation visualizations.'' These items were also rated on 5-point Likert scales.

Finally, we included one more way to assess trust, by asking participants whether they felt they trusted any of the four models enough that they would use them in a real deployment. Given the high-stakes nature of the task (loan approval), we felt that participants would only answer in the affirmative if they truly understood and trusted the pipelines produced by AutoAI.


\subsection{Results}
We begin our analysis by evaluating the reliability of our trust and understandability scales. Then, we examine the effect of the transparency features on trust and understanding of the AutoAI system~\cite{web:ibmautoai}.

When conducting analyses of variance, we controlled for the effects of gender, education level (undergraduate or graduate), prior experience with automated ML (used previously or not), and the first packet seen by including these terms in the model. In addition, when making comparisons between the V1 and V2 packets (a within-subjects factor), we include participant ID in the model as a random effect. We report effect sizes from our ANOVA models using partial $\eta^2$, which corresponds to the proportion of variance accounted for by each main effect, controlling for all other effects~\footnote{Miles \& Shevlin~\cite{miles2001applying} advise that a partial $\eta^2$ of $\geq .01$ corresponds to a small effect, $\geq .06$ to a medium effect, and $\geq .14$ to a large effect.}.

\subsubsection{Reliablity}
Factor analysis~\cite{thompson2004exploratory} indicated a high degree of reliability\footnote{Reliability indicates the extent to which items in the scale measure the same underlying conceptual construct. Common convention holds that $\alpha$ values greater than $0.70$ are considered reliable, and we refer to Tavakol and Dennick~\cite{tavakol2011making} for a more detailed discussion.} for the trust scale (Cronbach's $\alpha = 0.89$). Thus, despite our minor modifications to the wording of scale items, we see that its reliability falls within the $[0.87 - 0.92]$ range previously reported in Hoffman et al.~\cite{hoffman2018metrics}.

Analysis of the understandability scale also indicates a high level of reliability (Cronbach's $\alpha = 0.80$), without the need for dropping any items. Therefore, we construct two outcome measures for trust and understandability based on the averaged responses for each set of questions.

\subsubsection{Effect of Transparency Features on Trust and Understandability}
Overall ratings of trust for the V1 packet fell in the middle of the scale (M (SD) = 3.2 (.81) of 5), indicating that participants had neutral feelings about their trust of the base AutoML system. Ratings of understandability were higher (M (SD) = 3.6 (.52) of 5), indicating that they although they generally understood the information presented, there was also room for improvement.

We first compare the effect of having \emph{either} of the transparency features on ratings of trust and understandability. Participants had more trust in the V2 packets (M (SD) = 4.1 (.62)) than the V1 packets, and this difference was significant and large, $F [1,19] = 19.5$, $p < .001$, partial $\eta^2 = .36$. Prior experience with AutoML was also a marginally significant predictor of trust, $F [1,15] = 3.2$, $p = .09$, partial $\eta^2 = .06$. Participants with prior AutoML experience had more trust in both the V1 and V2 packets than participants without prior AutoML experience.

Participants had a greater understanding of the V2 packets (M (SD) = 4.0 (.54)) than the V1 packets, and this difference was also significant and large, $F [1,19] = 21.9$, $p < .001$, partial $\eta^2 = .49$. Unlike trust, participants with prior AutoML experience did not differ in their ratings of understandabilty, $F [1,15] = .17$, $p = n.s.$

We observe that the inclusion of either transparency feature caused a significant increase in ratings of trust and understandability over the base AutoML interface. We next seek to understand the extent to which each individual transparency feature -- input data distributions and feature engineering process -- affected ratings of trust and understandability. As these are between-subjects comparisons (participants only experienced one of these features), we no longer include participant ID as a random effect in our ANOVA model.

Ratings of trust for the input data distribution variant were higher (V2A M (SD) = 3.9 (.65)) than ratings of trust for the feature engineering process variant (V2B M (SD) = 4.2 (.60)), although this difference was not significant, $F [1,14] = .02$, $p = n.s.$ Ratings of understandability were equivalent for both variants (V2A M (SD) = 4.0 (.51), V2B M (SD) = 4.0 (.59), $F [1,14] = .12$, $p = n.s.$). Therefore, we conclude that the inclusion of \emph{any} transparency feature improved both trust and understandability, but the relative importance of either feature on improving trust and understandability remains unclear.

\subsubsection{Would Participants Deploy AutoML Models?}
The decision to deploy an AutoAI~\cite{web:ibmautoai} model was significantly correlated with both trust (Pearson's $r = .67$, $p < .001$) and understandability (Pearson's $r = .38$, $p = .02$). In general, participants did \emph{not} trust the V1 models enough for deployment: 17 participants (85\%) said they would not deploy any of the pipelines produced in V1. In contrast, participants trusted the V2 models more, with 16 participants (80\%) saying that they would deploy one of the pipelines produced in V2.

As with the previous results, we do not see clear differences between the V2A and V2B variants: 7 of 10 participants having V2A would deploy one of its pipelines, and 9 of 10 participants having V2B would deploy one of its pipelines. This difference was not significant, $\chi^2 = .31$, $p = n.s.$

We again find evidence that the inclusion of a transparency feature improved trust, but the relative importance of the two features we examined remains unclear.

\section{Study 3: Eliciting Information Needs}
\label{sec:study3}

In order to understand peoples' information needs in an AutoML user interface, we conducted a card-sorting exercise in which participants rank-ordered individual ``nuggets'' of information that might be included in an AutoML UI. This information was based on our examination of the kinds of information present across the AutoML interfaces of multiple vendors (discussed in Section~\ref{sec:study1}).

Each card took the form of a verb (i.e. ``see,'' ``visualize,'' ``know how'') followed by a piece of information related to the AutoML tool. For example, one nugget was ``view pre-processed data,'' and another was ``know how features are engineered.''


\subsection{Participants}
This study ran concurrently with Study 2. First, participants completed the task described in Section~\ref{sec:study2}. Next, they they completed the card-sorting task described below. The same set of participants completed both Study 2 and Study 3 in the same session.

\subsection{Procedure}
In this study, we used an open card-sorting method~\cite{spencer2004card, spencer2009card} to gain insight into what information is important for data scientists in order to trust the models produced by an AutoML tool. Participants were provided with 27 cards, each with a different nugget of information pertaining to an AutoML run based on our results from Study 1. Participants were also provided with blank cards to fill in additional information needs they identified. Participants were asked to sort the cards from ``most important for trust'' to ``least important for trust.''

\subsection{Results}
The card sorting exercise provided us with the opportunity to better understand the relative importance of different pieces of information that might be reported by AutoML. Although we performed an open card sorting task, in which participants were able to write in new informational requirements on blank cards, only four participants opted to do this. Therefore, we give an overall accounting of how cards were sorted by first ignoring these new cards, and then we provide detail on the content of these new cards.

We analyzed the card sorting results by computing the mean rank assigned to a card across participants. These ranks are shown in Table~\ref{tab:card_sort}. Each card is categorized in two ways: which aspect of the process it represents (process, data, or model), and whether the information is rendered as a visualization.

\begin{table}[htp]
    \small
    \centering
    \begin{tabular}{lllp{5cm}}
        \textbf{\#} & \textbf{Aspect} & \textbf{Transp.} & \textbf{Description} \\
        \hline
        \multicolumn{3}{l}{\emph{Most important for trust}} \\
        26 & E   & Model   & View evaluation metrics                                      \\
        27 & E   & Model   & View visualizations of model performance                     \\
        10 & PPD & Process & Know how raw data was pre-processed                          \\
        2  & RD  & Data    & View the meanings of each column in the raw data             \\
        4  & RD  & Data    & Visualize each column's distribution in the raw data         \\
        5  & RD  & Data    & Visualize the raw data - view overall distributions          \\
        23 & P   & Process & View process of how a pipeline is created                    \\
        7  & RD  & Data    & View the raw data - statistics of individual distributions   \\
        3  & RD  & Data    & Visualize outliers in the raw data                           \\
        8  & RD  & Data    & View statistics of missing values in the raw data            \\
        11 & PPD & Data    & View statistics of the pre-processed data                    \\
        17 & FE  & Model   & Effect of engineered features                                \\
        12 & PPD & Data    & Visualize data after pre-processing                          \\
        19 & P   & Model   & Show adopted models in output pipelines                      \\
        6  & RD  & Data    & View statistics of outliers in raw data                      \\
        15 & FE  & Data    & View how existing features were engineered into new features \\
        24 & P   & Model   & Ability to edit a pipeline                                   \\
        25 & E   & Model   & Compare differences between pipelines                        \\
        1  & RD  & Data    & View the raw data table                                      \\
        18 & P   & Process & Show which types of models considered for model selection    \\
        9  & PPD & Data    & View the pre-processed data table                            \\
        20 & E   & Model   & Compare one model against other models                       \\
        14 & FE  & Model   & View new engineered features                                 \\
        13 & RD  & Data    & See how data was split (test vs. train/holdout)              \\
        16 & FE  & Process & Know how features were engineered                            \\
        21 & HP  & Model   & See model's hyperparameters                                  \\
        22 & HP  & Process & Know how hyperparameter optimization was performed           \\
        \multicolumn{3}{l}{\emph{Least important for trust}} \\
        \hline
    \end{tabular}
    \caption{Ranking of importance of different kinds of information in an AutoML user interface. Items at the top of the list were rated as being more important for establishing trust in an AutoML system. Informational aspects are: \emph{RD}: raw data, \emph{PPD}: pre-processed data, \emph{HP}: hyperparameters, \emph{P}: pipeline, \emph{E}: model evaluation, \emph{FE}: feature engineering.}
    \label{tab:card_sort}
\end{table}

\subsubsection{Information Needs for Establishing Trust}
Perhaps not surprisingly, information pertaining to the performance of the generated pipelines was rated as the most important for trust, either as a raw metric (\#26) or in visual form (\#27). Also important were several aspects related to process: knowing how data were pre-processed before training (\#10) and being able to view the process by which a pipeline was created (\#23). Other process-related information was deemed less important, such as knowing which types of models were considered for model selection (\#18) and knowing how hyperparameter optimization was performed (\#22).

In our analysis of the Study 2 results, we did not see a significant difference in ratings of trust between participants who were given visualizations of input data distributions (\#5) and participants who were given information on how feature engineering worked (\#16). However, from the card-sorting exercise, we see a clear difference in how participants ranked the importance of these two features: participants felt that being able to see input data distributions was more important.

We were struck by participants' relative lack of interest in feature engineering (FE in Table~\ref{tab:card_sort}, mean rank of 19/27). Participants were more interested in the raw data (mean rank of 11/27) and the model evaluation metrics (mean rank of 11/27). However, in the course of data science work, raw data are transformed and engineered into features. Therefore, data scientists' relative lack of interest in the design of these features~\cite{feinberg, muller} should be examined in future research.


\section{Discussion}
The goal of our research is to explore trust in the relationship between human data scientists and AutoML systems. Based on previous literature \cite{hancock2011meta, wang2019atmseer,siau2018building}, we were interested in how the inclusion of transparency features in an AutoML system affects user trust and understanding of the tool (RQ1) and identifying the most important information data scientists need to establish trust in an AutoML system (RQ2). 

We find that including certain transparency features such as visualizations of input data distributions and a graphic depicting the feature engineering process does improve user trust and understandability of AutoML. The inclusion of either transparency feature had a significant and sizable effect on both trust and understandability. Although we were unable to uncover a statistical difference between the two transparency features in Study 2, results from the card-sorting exercise (Study 3) showed users clearly ranked the ``input data distributions'' feature more important. Our findings support the hypothesis that the increase of transparency through adding additional information significantly lead to the increased trust and understandability. Thus, we provide support for RQ1, and we propose that future work should use a more sensitive, full repeated-measures experimental design, to clarify RQ2.


\subsection{Implications for Design and Future Research}

\subsubsection{Visualizations}

Our studies illustrate that conversations with the input data \cite{feinberg, muller} and information on the processes of feature engineering are valuable. Future work should explore the effectiveness of different types of data visualizations for both input data and also post-feature-engineered data. How do users want to examine these data? Are there mini-visualizations that can be applied "in place" in a pipeline diagram? What are effective ways to "sample" the data at different points in a pipeline. Will it be possible to recover transformations or feature-engineering algorithms from past analyses \cite{kery2019towards} for comparison with current data and features?

If users will be sampling data along a pipeline, then we may also want to explore effective \textit{comparative} visualizations. Similarly to the ManyEyes project \cite{viegas} with data analysts, we may also want to explore data scientists' needs for annotation and communication regarding data visualizations along a pipeline. What individual problems do data scientists want to solve in this way? What collaborative problems do they want to address collectively, and what types of messages would be helpful?

\subsubsection{Individual Differences and Personalization}

We recognize that there are many individual differences across those who practice data science \cite{arya2019one, kross2019practitioners}: difference in backgrounds including background knowledge, skills, work practices, and experience levels make it difficult to claim that AutoML tools ought to be designed as ``one size fits all'' \cite{hou2017hacking}. We found a wide range of individual difference and preferences, even within our (relatively) small sample of 24 participants. For example, while thinking aloud during Study 3, six participants explicitly stated that information pertaining to the raw data was most important for them. In contrast, three participants said that the processed data was more important than the raw data, and two participants commented that the raw data was the least important and they had no desire to see this information in the tool. These differences in individual preferences can create more complications when we consider the fact that domain experts and data science workers need to work closely together, as suggested in\cite{hou2017hacking,mao2019}. 

Our findings suggest that AutoML tools may need to allow for a degree of personalization to accommodate individual preferences or different domains of use. Recent research by Arya et al.~\cite{arya2019one} addresses these concerns by defining explanation methods for different audiences and domains. 

\subsubsection{Context of Use}
In all three of our studies, participants identified a dichotomy between using AutoML for research purposes and using it in their day-to-day work practice. From these discussions, we recognized that the importance of different kinds of information depends on the intended use of the tool as well. For example, in Study 3, the two cards relating to hyperparameter optimization (\#21, \#22) had the lowest mean rank. But we should not conclude that this information is therefore unimportant. Further qualitative interview confirmed our speculation: two participants explicitly mentioned that information about hyperparameter optimization would be more important if the scenario of use was focused on conducting research rather than building models to approve loans.

In addition, we discovered through discussions with our participants that there is a difference between trust in a model produced by AutoML and trust in an AutoML tool itself. One participant in Study 2 commented that even though they may produce a model with low accuracy and not wish to deploy it, they would still maintain their trust in the AutoML tool itself. This disparity of trusting AutoML's artifacts versus trusting the AutoML itself is one research topic that ought to be explored further.

\subsection{Limitations}
There are several limitations of our work that may limit our ability to draw broad conclusions from our findings. First, our participants were drawn from a pool of undergraduate and graduate students having prior experience in data science work. However, their experience does not necessarily generalize to that of professional data scientists, whose information needs for establishing trust in AutoML may differ. Given the preponderance of AutoML systems being developed by and for enterprise users (e.g.~\cite{web:ibmautoai, web:microsoftazure, web:h2o, web:googleautoml, web:datarobot}), additional work is needed to examine the viewpoints of professional data scientists. As a future work, we plan to validate our results with different user groups.

The dataset used in this study is a widely-used loan application in data science training~\cite{web:kaggle}. This application domain, together with some other domains (e.g., healthcare and jurisdiction), are crucially important to the involved individuals, families, and businesses. Recent literature~\cite{hajian2016algorithmic,van2017responsible} have suggested some of the datasets or AI algorithms may inherit the discrimination against people of color, or women, or women of color, in approving loans. Future work can further investigate transparency features of presenting bias-detection and bias-mitigation information as part of the initial trust establishment.

It should also be noted that the card sorting task performed by the participants who are first-time users of AutoML system generated information needed for establishing trust and may not represent the needs of data scientists who have an established relationship with an AutoML system. The literature suggests that trust formation and trust retention are different and therefore they require different considerations in AutoML systems \cite{siau2018building,6468022}.

In addition, during Study 3, participants were asked to sort the cards on a scale of most important to least important for establishing trust. As participants thought aloud about their sorting decisions, some noted that the ranks of the cards could change over time as they continued to interact with the tool. Therefore, the results of the card rankings may only represent the information needed to \emph{establish initial trust}, which may be different from information needed to maintain trust.

\section{Conclusion}
As the AI industry is expected to grow significantly in the next few years \cite{carvalho2019machine}, research on the relationship between user trust and automated data science systems is critical to ensure these tools can be trustworthy enough to be adopted responsibly by the public. Transparency is known to be a significant factor in trusting automated data science systems. However, much of the literature states a lack of transparency in AI \cite{Weitz:2019:YTM:3308532.3329441,carvalho2019machine,Yin:2019:UEA:3290605.3300509}. We believe exploring the information needs and individual differences of data scientists can inform us of possible ways to increase trust in AutoML tools. Therefore, in this work we showed that increasing transparency via providing a user with more information about an AutoML tool significantly increased user trust as well as user understandability in the tool. By gathering the information requirements of data scientists to establish trust in these tools we have provided a pool of transparency features that can each be further researched to see how they impact user trust. As our work suggests it may be unreasonable to design an AutoML tool suitable for all users across all domains, we encourage the data science, HCI, and human studies research communities to continue exploring how to accommodate different AutoML users and enhance their trust in the tools based on their domains, knowledge, and common practices.



\bibliographystyle{ACM-Reference-Format}
\bibliography{main}

\end{document}